\documentclass[10pt,journal,letterpaper,final,compsoc,twoside]{IEEEtran}
\usepackage[nocompress]{cite}
\usepackage[pdftex]{graphicx}
\usepackage[cmex10]{amsmath}
\usepackage{amssymb,amsthm}
\usepackage[T1]{fontenc}
\usepackage[utf8]{inputenc}
\theoremstyle{remark}
\newtheorem{Remark}{Remark}
\theoremstyle{plain}
\newtheorem{Theorem}{Theorem}
\newcommand{\koerper}[1]{\mathbb{#1}}
\newcommand{\R}{\koerper{R}}

\providecommand{\abs}[1]{\lvert#1\rvert}

\providecommand{\mvec}[1]{\mathbf{#1}}

\begin{document}

\title{A class of random fields on complete graphs with tractable partition
function}
\author{Boris~Flach,~\IEEEmembership{Member,~IEEE}%
\thanks{B.~Flach is with Czech Technical University in Prague, Czech Republic}}

\markboth{Digital Object Indentifier 10.1109/TPAMI.2013.99
\hfil 0162-8828/13/\$31.00 © 2013 IEEE}{This article has been accepted for
publication in a future issue of this journal, but has not been fully edited.}
\IEEEcompsoctitleabstractindextext{%
\begin{abstract}
The aim of this short note is to draw attention to a method by which the
partition function and marginal probabilities for a certain class of random
fields on complete graphs can be computed in polynomial time. This class
includes Ising models with homogeneous pairwise potentials but arbitrary
(inhomogeneous) unary potentials. Similarly, the partition function and marginal
probabilities can be computed in polynomial time for random fields on complete
bipartite graphs, provided they have homogeneous pairwise potentials. We expect
that these tractable classes of large scale random fields can be very useful for
the evaluation of approximation algorithms by providing exact error estimates.
\end{abstract}
}

\maketitle
\section{Introduction}
The computation of unary and pairwise marginals for probabilistic graphical
models is necessary both for inference and learning if these models are used in
pattern recognition and computer vision. It is well known that this problem is
\#P-hard for Markov/Gibbs Random Fields if there are no restrictions for the 
underlying graph structure of the model\cite{Bulatov:TCS2005}. This explains the
strong focus on approximation algorithms for the calculation of marginal
probabilities. 

Ising like models on complete but {\em small} graphs are often considered in
error estimation experiments, when proposing new or improved methods for
approximate calculation of marginal probabilities
\cite{Wainwright:TIT2005,Wainwright:TSP2006,Sontag:NIPS07,
Hazan:TIT2010,Korc:ICML2012}. We show in this note that the partition function
and marginal probabilities for a certain class of random fields on complete
graphs can be computed in polynomial time. This class includes Ising models with
homogeneous pairwise potentials but arbitrary (inhomogeneous) unary potentials.
Similarly, the partition function and marginal probabilities can be computed in
polynomial time for random fields on complete bipartite graphs, provided they
have homogeneous pairwise potentials.

The main idea is to partition the set of all labellings so that the following
holds for each subset: (i) the contribution of the pairwise factors is equal for
all labellings in a subset, (ii) the sum of contributions of the unary factors
can be computed by dynamic programming over the graph size. This results, e.g.,
in an algorithm with $\mathcal{O}(n^K)$ time complexity for computing the
partition sum for $K$-valued random fields on a complete graph with $n$
vertices, provided that the pairwise factors of the GRF are homogeneous. 
\section{The model class; computing the partition function}
Let us consider the following class of binary valued random fields on
undirected complete graphs
\begin{equation}\label{eq:model}
 p(x) = \frac{1}{Z} 
 \prod_{\{ij\}\in E} g(x_i,x_j) \prod_{i\in V} q_i(x_i) ,
\end{equation}
where $(V,E)$ denote the sets of vertices and edges of a complete graph and
$x\colon V\rightarrow \{0,1\}$ is a binary valued labelling of the vertices.
Notice that we assume that $g\colon \{0,1\}^2\rightarrow \R$ is shared by all
pairwise factors. Given the model parameters $g$ and $q_i$, the task is to
compute the partition sum
\begin{equation}\label{eq:partition_sum}
 Z = \sum_{x\in \mathcal{X}}
 \prod_{\{ij\}\in E} g(x_i,x_j) \prod_{i\in V} q_i(x_i) 
\end{equation}
as well as unary and pairwise marginal probabilities $p(x_i)$ and $p(x_i,x_j)$.
We assume without loss of generality that $g$ has the form
\begin{equation}\label{eq:pairwise pots}
 g(k,k') = \begin{cases}
            \alpha & \text{if $k\neq k'$,} \\
            1 & \text{otherwise,}
           \end{cases}
\end{equation}
and the unary factors have the form
\begin{equation}\label{eq:unary pots}
 q_i(k) = \begin{cases}
            \beta_i & \text{if $k=0$,} \\
            1 & \text{otherwise.}
           \end{cases}
\end{equation}
This can be achieved by applying an appropriate re-parametrisation without
changing the probability \eqref{eq:model}.

In order to calculate the partition sum $Z$ for a graph with $n$ vertices, we
partition the set $\mathcal{X} =\{0,1\}^n$ of all labellings  into the sets
$\mathcal{X}_0,\mathcal{X}_1,\ldots,\mathcal{X}_n$, where 
\begin{equation}
 \mathcal{X}_m = \Bigl\{
 x\in\mathcal{X} \Bigm | \sum_{i\in V} x_i = m \Bigr\}  
\end{equation}
denotes the set of all labellings with $m$ vertices labelled by ``$1$''.
Accordingly, we denote the partial sums by 
\begin{equation} \label{eq:partial_sum}
 Z_m = \sum_{x\in \mathcal{X}_m}
 \prod_{\{ij\}\in E} g(x_i,x_j) \prod_{i\in V} q_i(x_i) .
\end{equation}
Due to the homogeneity assumption \eqref{eq:pairwise pots}, the pairwise factors
in \eqref{eq:partial_sum} contribute to each summand of $Z_m$ by the same factor
$\alpha^{m(n-m)}$. Hence, we can write
\begin{equation}
 Z = \sum_{m=0}^n Z_m = \sum_{m=0}^n \alpha^{m(n-m)} H_V(m),
\end{equation} \label{eq:unary_contrib}
where 
\begin{equation}
 H_V(m) = \sum_{x\in\mathcal{X}_m} \prod_{i\in V} q_i(x_i) = 
 \sum_{x\in\mathcal{X}_m} \prod_{i\in V} \beta_i^{1-x_i}
\end{equation}
denotes the sum of all unary contributions to the partial sum $Z_m$.
These quantities can be computed recursively over the size of the graph. Let us
denote by $H_U(m)$, $m=0,1,\ldots,\abs{U}$ the corresponding quantities for the
complete sub-graph induced by the vertex set $U\subset V$. If $i\in U$ is a
vertex of this graph, then 
\begin{equation} \label{eq:recur_map}
 H_U(m) = \beta_i \, H_{U\setminus i} (m) + H_{U\setminus i}(m-1) .
\end{equation}
This equation follows from the simple observation that any labelling of $U$ with
$m$ vertices labelled by ``1'' either has the vertex $i$ labelled by ``1'' and
$m-1$ of the remaining vertices labelled by ``1'' or has the vertex $i$ labelled
by ``0'' and consequently $m$ of the remaining vertices labelled by ``1''.
Hence, the quantities $H_V(m)$ can be computed by dynamic programming over the
size of the graph, what eventually leads to an algorithm for computing $Z$ with
$\mathcal{O}(n^2)$ time complexity.

It is similarly easy to compute marginal probabilities because the mapping
defined by \eqref{eq:recur_map} is invertible. In order to compute
e.g.~the unary marginal probabilities for the vertex $i\in V$
\begin{align}
 p(x_i=1) &\propto \sum_{m=1}^n 
 H_{V\setminus i}(m-1) \; \alpha^{m(n-m)} \\
  p(x_i=0) &\propto \beta_i \sum_{m=0}^{n-1}
 H_{V\setminus i}(m) \; \alpha^{m(n-m)} ,
\end{align}
we need the quantities $H_{V\setminus i}(m)$, which can be computed from those
for the whole graph with $\mathcal{O}(n)$ time complexity by
\begin{equation} \label{eq:recur_map_inv}
 H_{V\setminus i} (m) = \frac{1}{\beta_i} \bigl[
 H_V(m) - H_{V\setminus i} (m - 1) \bigr] .
\end{equation}
This results in an $\mathcal{O}(n^2)$ algorithm for calculating the unary
marginals for {\em all} vertices of the graph.

The proposed approach can be generalised to $K$-valued random fields on complete
graphs with homogeneous pairwise interactions. The probability and partition
function are given by \eqref{eq:model} and \eqref{eq:partition_sum} as before,
but $x$ denotes now a realisation of a field of $K$-valued random variables.
Notice that the pairwise factors $g(\cdot,\cdot)$ are assumed to be symmetric up
to a re-parametrisation. We partition the set of all labellings $\mathcal{X}$
into sets $\mathcal{X}_{\mvec{m}}$, where $\mvec{m} = (m_1,\ldots,m_K)$ is a
vector, the components of which denote the number of variables taking the
respective label value, i.e,
\begin{multline}
 \mathcal{X}_{\mathbf{m}} = \\
 \Bigl\{x\in \mathcal{X} \Bigm |
 \sum_{i\in V} \delta_{x_i 1} = m_1 \; , \ldots , \;
 \sum_{i\in V} \delta_{x_i \scriptstyle K} = m_K
 \Bigr\} ,
\end{multline}
where $\delta_{ij}$ denotes the Kronecker delta.
The corresponding contributions $H_V(\mathbf{m})$ of all unary terms to the
partial sums can be again computed recursively by
\begin{equation}
 H_U(\mvec{m}) = \sum_{k=1}^K q_i(k) \;
 H_{U\setminus i} \bigl(\mvec{m} -\mvec{e}_k \bigr) ,
\end{equation}
where $\mvec{e}_k$ denotes the standard basis vector for the component $k$.
Since there are $\mathcal{O}(n^{K-1})$ quantities $H_V(\mathbf{m})$ and each of
them must be recomputed $n$ times, the overall time complexity for computing
them is $\mathcal{O}(n^K)$. Finally, the partition sum is obtained using the
pairwise factors $g(\cdot,\cdot)$
\begin{equation}
 Z = \sum_{\mvec{m}} H_V(\mathbf{m}) 
 \prod_{k\leqslant k'} g(k,k')^{n(m_k,m_{k'})} ,
\end{equation}
where 
\begin{equation}
 n(m_k,m_{k'}) = 
 \begin{cases}
  m_k m_{k'} & \text{if $k\neq k'$,} \\
  m_k (m_k -1 ) / 2 & \text{otherwise}
 \end{cases}
\end{equation}
is the number of edges connecting vertices labelled by $k$ and $k'$ for
labellings in $\mathcal{X}_{\mvec{m}}$. 

In order to compute all unary marginal probabilities 
\begin{multline}
 p(x_i = k) \propto  q_i(k) \times \\
  \sum_{\mvec{m}} 
 H_{V\setminus i} (\mvec{m}) 
 \prod_{k'\leqslant k''} g(k',k'')^{\widetilde{n}(m_{k'},m_{k''})} ,
\end{multline}
it is necessary to compute the quantities $H_{V\setminus i} (\mvec{m})$ for all
$i\in V$ and all $\mvec{m}$. This can be done efficiently because the mapping
$H_{V\setminus i} (\cdot) \mapsto H_{V} (\cdot)$ is invertible as
before in the case of binary valued random fields. Hence, each of the
$\mathcal{O}(n^K)$ quantities $H_{V\setminus i} (\mvec{m})$ can be computed in
constant time from the previously computed quantities $H_{V}(\mvec{m})$.

Altogether this leads to the following theorem.
\begin{Theorem}
Suppose a $K$-valued random field defined on a complete graph with $n$ vertices
has homogeneous pairwise factors and arbitrary (inhomogeneous) unary factors.
Then its partition sum as well as all its unary marginals can be computed with
$\mathcal{O}(n^K)$ time complexity.
\end{Theorem}
\begin{Remark}
Note that this theorem is not in contradiction with the dichotomy found by
Bulatov and Grohe \cite{Bulatov:TCS2005} because here we restrict the graph
structure.
\end{Remark}
\section{Models on complete bipartite graphs}
The applicability of the proposed approach can be extended even further,
e.g., for the calculation of the partition function and marginal probabilities
for random fields defined on complete bipartite graphs. Here, again, it is
required that the pairwise interactions are homogeneous. We will consider binary
valued random fields for the sake of simplicity.

Let $G(A\cup B,E)$ be a complete bipartite graph such that $n_1=\abs{A}$,
$n_2=\abs{B}$. Consider the class of binary valued random fields $x\colon
A\cup B\rightarrow \{0,1\}$
\begin{equation} \label{eq:model_bip}
 p(x) = \frac{1}{Z} 
 \prod_{\substack{i\in A\\ j\in B}} g(x_i,x_j) \prod_{i\in A\cup B} q_i(x_i) .
\end{equation}
As discussed in the previous section, we may assume that the factors $g$, $q_i$
have the form \eqref{eq:pairwise pots} and \eqref{eq:unary pots} respectively.

In order to compute $Z$, we partition the set of all labellings $\mathcal{X}$
into the sets 
\begin{equation}
  \mathcal{X}_\mvec{m} = \Bigl\{
 x\in\mathcal{X} \Bigm | \sum_{i\in A} x_i = m_1 \text{ , }
 \sum_{j\in B} x_j = m_2 \Bigr\} ,
\end{equation}
where $\mvec{m}=(m_1,m_2)$ counts the number of vertices labelled by ``1'' in
the first and second part, respectively. It is clear that the pairwise factors
contribute to all summands of $Z_\mvec{m}$ by the same factor $\alpha^\kappa$,
where $\kappa=m_1(n_2 - m_2) + m_2 (n_1 - m_1)$. Hence, as before,
the problem
reduces to the computation of the contributions
\begin{equation}
 H_{AB}(\mvec{m}) = 
 \sum_{x\in\mathcal{X}_\mvec{m}} \prod_{i\in A\cup B} \beta_i^{1-x_i}
\end{equation}
of the
unary factors. They can be computed recursively over the graph size. Similar to
\eqref{eq:recur_map} we have here
\begin{align}
 H_{U V}(\mvec{m}) &= \beta_i \cdot 
 H_{U\setminus i\;V}(\mvec{m}) + 
 H_{U\setminus i\;V}(\mvec{m} - \mvec{e}_1) ,\label{eq:recur_map1}\\
 H_{U V}(\mvec{m}) &= \beta_j \cdot 
 H_{U V\setminus j}(\mvec{m}) + 
 H_{U V\setminus j}(\mvec{m} - \mvec{e}_2) . \label{eq:recur_map2}
\end{align}
This results in an $\mathcal{O}(n_1^2 n_2^2)$ algorithm for computing the
partition sum $Z$. Both mappings \eqref{eq:recur_map1}, \eqref{eq:recur_map2}
are invertible, which yields an algorithm for calculating all unary marginal
probabilities with the same time complexity. Generalising this to $K$-valued
labellings as in the previous section yields the following result.
\begin{Theorem}
Suppose a $K$-valued random field defined on a complete bipartite graph with
$n_1+n_2$ vertices has homogeneous pairwise factors and arbitrary
(inhomogeneous) unary factors. Then its partition sum as well as all its unary
marginals can be computed with $\mathcal{O}(n_1^K n_2^K)$ time complexity.
\end{Theorem}
\section{Conclusion} 
We have shown that the partition sum and marginal probabilities can be
efficiently computed for random fields on complete graphs if they have
homogeneous pairwise factors. Similarly, the partition sum and marginal
probabilities can be efficiently computed for random fields on complete
bipartite graphs, provided they have homogeneous pairwise factors. We do not
expect these two model classes to be directly relevant for computer vision
applications. We expect, however, that they can be very useful to evaluate
approximation algorithms for computing marginal probabilities. To the best of
our knowledge, they are the only known classes of random fields on graphs with
large tree-width and with arbitrary unary factors\footnote{The method proposed
in \cite{Schraudolph:NIPS2009} requires outer-planar graphs, i.e.~tree-width
two, if applied for the case of arbitrary unary factors.} for which the marginal
probabilities can be computed in polynomial time, thus providing exact error
estimates for approximation algorithms.
\section*{Acknowledgements}
\addcontentsline{toc}{section}{Acknowledgements}
The author has been supported by the Grant Agency of the Czech Republic, project
P202/12/2071. He would like to thank the reviewers, especially for the
exceptionally thorough and favourable consideration that greatly helped to
improve the original manuscript. The author wants to express his special
gratitude to Tomas Werner for valuable discussions, which were as usual
strict {\em and} motivating.
\bibliographystyle{IEEEtran}
\bibliography{IEEEabrv,npf-2012}
\begin{IEEEbiography}[{\includegraphics[width=1in,height=1.25in,clip,
keepaspectratio]{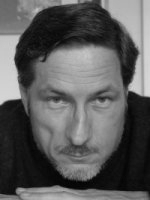}}]{Boris Flach}
is an associate professor at the Czech Technical University in Prague, Center
for Machine Perception. He received the PhD degree in theoretical physics (1989)
and Habilitation in computer science (2003) from Dresden University of
Technology. His main research interests are in probabilistic graphical models
and statistical machine learning.
\end{IEEEbiography}
\vfill
\end{document}